\documentclass[11pt]{article}

\usepackage[a4paper,top=22mm,bottom=25mm,left=25mm,right=25mm]{geometry}
\usepackage[tt=false]{libertine}
\usepackage[libertine]{newtxmath}
\usepackage[scaled=0.92]{zi4}
\usepackage[semibold]{sourcesanspro}
\newcommand{\DFBrandFont}{\sffamily\bfseries}
\newcommand{\DFTitleFont}{\sffamily\bfseries}
\usepackage{microtype}
\usepackage{graphicx}
\usepackage[table]{xcolor}
\usepackage{booktabs}
\usepackage{array}
\usepackage{tabularx}
\usepackage{tcolorbox}
\usepackage{caption}
\usepackage{enumitem}
\usepackage{amsmath}
\usepackage{pifont}
\usepackage[numbers,sort&compress]{natbib}
\usepackage{hyperref}
\usepackage{fancyhdr}
\usepackage{lastpage}
\usepackage{float}
\usepackage{placeins}

\definecolor{DFOrange}{HTML}{F97316}
\definecolor{DFOrangeDark}{HTML}{EA580C}
\definecolor{DFOrangeSoft}{HTML}{FFF3E8}
\definecolor{DFOrangeLine}{HTML}{FDBA74}
\definecolor{DFInk}{HTML}{111827}
\definecolor{DFMuted}{HTML}{4B5563}
\definecolor{DFTable}{HTML}{F8FAFC}
\definecolor{DFOurs}{HTML}{FFE2CC}
\definecolor{DFOursDark}{HTML}{C2410C}

\hypersetup{
  colorlinks=true,
  linkcolor=DFOrangeDark,
  citecolor=DFOrangeDark,
  urlcolor=DFOrangeDark,
  pdftitle={DreamForge-World 0.1 Preview: A Low-Compute Real-Time Controllable World Model},
  pdfauthor={Daniyel Ayupov and Artur Markov-Tsoy}
}

\setlist[itemize]{leftmargin=*,topsep=3pt,itemsep=1pt,parsep=0pt}

\newcommand{\dfworld}{DreamForge-World 0.1 Preview}
\newcommand{\dfshort}{DF-World 0.1 Preview}
\newcommand{\lab}{DreamForge AI Lab}
\newcommand{\cmark}{\ding{51}}
\newcommand{\xmark}{\ding{55}}

\newcommand{\kmp}{keyboard/mouse}

\begin{document}
\thispagestyle{empty}

\noindent\begin{minipage}[t]{0.55\textwidth}
  \vspace{0pt}\includegraphics[height=0.44cm]{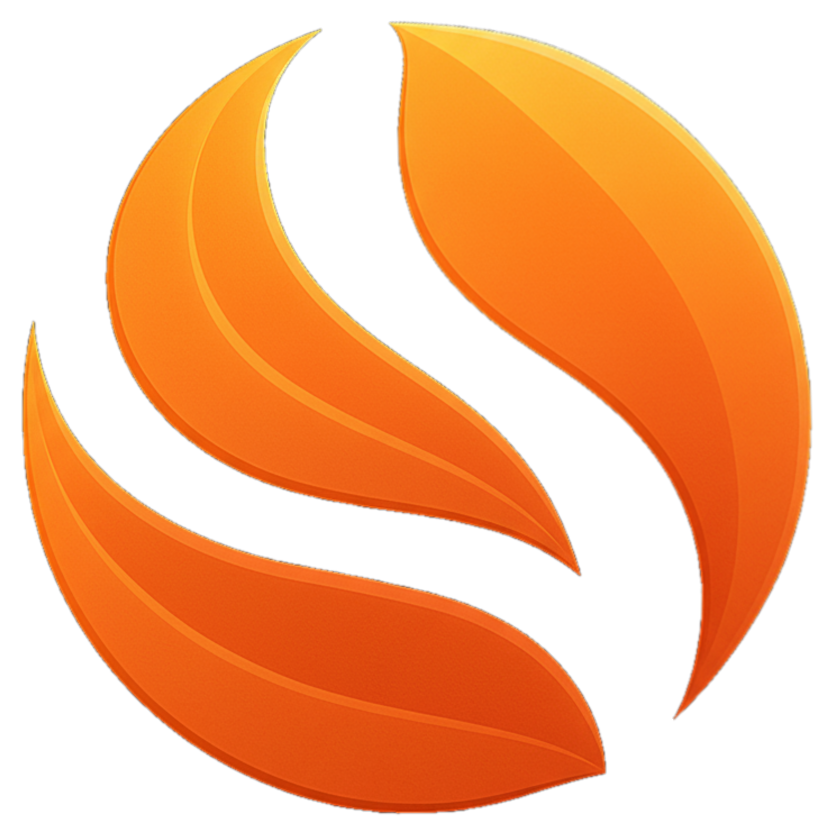}\hspace{0.36em}%
  \raisebox{0.095cm}{\DFBrandFont\fontsize{9.9}{12.2}\selectfont DreamForge AI Lab}
\end{minipage}%
\hfill
\begin{minipage}[t]{0.34\textwidth}
  \vspace{0.06cm}\raggedleft\small\textcolor{DFMuted}{Technical Report\\June 2026}
\end{minipage}

\vspace{0.45em}
\noindent\textcolor{DFOrange}{\rule{\textwidth}{0.55pt}}

\newlength{\DFTitleLineWidth}
\settowidth{\DFTitleLineWidth}{{\small \textsuperscript{1}\lab, Kazakhstan \quad $*$Corresponding author: ayupovdaniel07@gmail.com}}

\begin{center}
  \vspace{1.05em}
  {\resizebox{\DFTitleLineWidth}{\height}{\DFTitleFont\fontsize{23.4}{28.2}\selectfont \dfworld}}\par
  \vspace{0.34em}
  {\resizebox{\DFTitleLineWidth}{\height}{\DFTitleFont\fontsize{15}{16.0}\selectfont A Low-Compute Real-Time Controllable World Model}}\par
  \vspace{1.05em}
  {\large Daniyel Ayupov\textsuperscript{1} \quad Artur Markov-Tsoy\textsuperscript{1}}\par
  \vspace{0.25em}
  {\small \textsuperscript{1}\lab, Kazakhstan \quad Contact: \href{mailto:contact@updates.trydreamforge.com}{contact@updates.trydreamforge.com}}
\end{center}

\vspace{0.6em}

\begin{figure}[H]
  \centering
  \includegraphics[width=\textwidth]{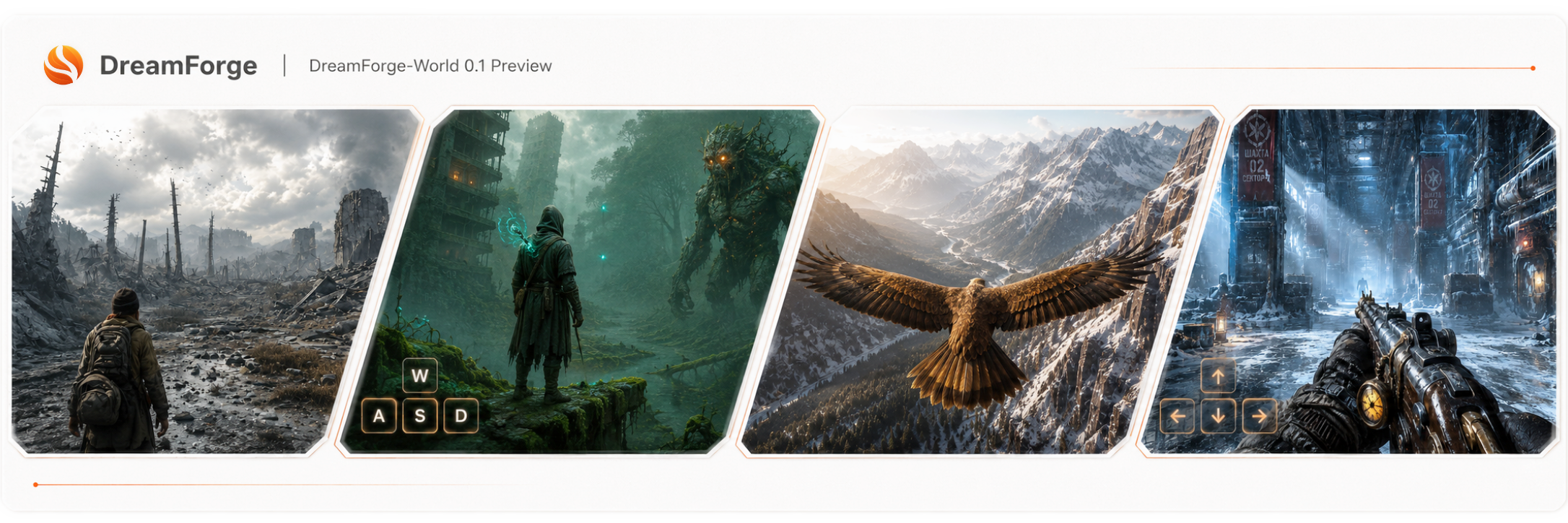}
  \caption{Representative DF-World 0.1 Preview domains and control overlays used in the first-page teaser.}
  \label{fig:banner}
\end{figure}

\begin{tcolorbox}[
  colback=DFOrangeSoft,
  colframe=DFOrangeLine,
  boxrule=0.55pt,
  arc=7pt,
  left=12pt,
  right=12pt,
  top=11pt,
  bottom=11pt]
\textbf{Abstract.} We present \dfworld{}, a preview foundational world model for real-time interactive world simulation. The system adapts the LongLive~1 autoregressive video stack, itself derived from Wan2.1-T2V-1.3B, with a residual action pathway inspired by the Matrix-Game family~\cite{wan2025,longlive2025,matrixgame2}. \dfworld{} focuses on a complementary axis to frontier-scale world simulators: low-compute adaptation, consumer-GPU runtime, and broad interactive capability coverage. It supports live keyboard and mouse control, multimodal initialization, mid-stream reprompting, dual-view operation, and minute-scale interactive rollouts at native 480p resolution, reaching up to 14--15~FPS on a single RTX~4090 with a low memory footprint. By leveraging open video backbones and applying targeted adaptation runs, we build the preview system with high cost-efficiency. \dfshort{} is not yet a memory-complete or frontier-quality world simulator, but demonstrates a practical low-compute route toward real-time controllable world-model previews on consumer GPUs.

\vspace{0.45em}
\textbf{Website:} \href{https://trydreamforge.com}{trydreamforge.com}
\end{tcolorbox}

\section{Introduction}

Interactive world models generate future visual observations in a closed control loop conditioned on visual history, user actions, and optional semantic inputs. Unlike offline video generation, the model must update the scene frame by frame under streaming control, while its own generated outputs become the next conditioning history. This shifts the problem from producing a plausible clip to maintaining controllable rollout under self-generated context, a framing shared by recent interactive generators and world simulators such as Genie, GameNGen, Matrix-Game, WorldPlay, LongLive, and DreamX-World~\cite{ha2018worldmodels,genie2024,gamengen2024,matrixgame2,worldplay,longlive2025,dreamxworld}.

The field is increasingly organized around a latency-versus-state-retention tradeoff. Real-time interaction benefits from short context windows, cacheable causal structure, few-step inference, quantization, and lightweight decoding. Long-horizon consistency instead benefits from memory retrieval, revisit-aware conditioning, camera-aware state, and training or distillation under self-generated histories~\cite{longlive2025,matrixgame2,matrixgame3,worldplay,infiniteworld,dreamxworld}. Systems such as Matrix-Game~3.0, WorldPlay, Infinite-World, DreamX-World, Genie~2, and Genie~3 therefore emphasize persistent or retrievable memory as a central component of interactive simulation~\cite{matrixgame3,worldplay,infiniteworld,dreamxworld,genie2,genie3}.

\dfshort{} targets a different operating point: broad interactive capability coverage under tight adaptation compute. The preview adapts the LongLive/Wan video-prior lineage into a streaming action-conditioned rollout system, then exposes text, image, video, and mixed initialization, mid-stream reprompting, and separate first- and third-person control paths.

The main contribution is a compact system recipe: adapt an open causal video prior, add a residual action pathway, seed the autoregressive history with multimodal inputs, and optimize the resulting loop for single-GPU preview interaction. This places \dfshort{} in the same design space as recent real-time world models, but on a complementary axis: capability coverage per unit direct adaptation compute rather than frontier visual fidelity, persistent spatial memory, or engine-level control precision.

\begin{table}[H]
  \centering
  \small
  \renewcommand{\arraystretch}{1.13}
  \setlength{\tabcolsep}{3.3pt}
  \rowcolors{2}{white}{DFTable}
  \resizebox{\textwidth}{!}{%
  \begin{tabular}{lccccccc}
    \toprule
    & Matrix-Game 2.0 & Matrix-Game 3.0 & HY-WorldPlay 1.5 & Waypoint 1.5 & Genie 3 & LingBot-World & \cellcolor{DFOrange}\textcolor{white}{\textbf{Ours}} \\
    \midrule
    Real-time on 1 GPU & \cmark & \cmark & \cmark & \cmark & \xmark & \xmark & \cellcolor{DFOurs}\cmark \\
    Memory & \xmark & \cmark & \cmark & \xmark & \cmark & \cmark & \cellcolor{DFOurs}\xmark \\
    Reprompting & \xmark & \xmark & \cmark & \cmark & \cmark & \cmark & \cellcolor{DFOurs}\cmark \\
    Diverse multimodal input & \xmark & \xmark & \xmark & \xmark & \xmark & \xmark & \cellcolor{DFOurs}\cmark \\
    Dual-view support & \xmark & \xmark & \cmark & \xmark & \cmark & \cmark & \cellcolor{DFOurs}\cmark \\
    Resolution & 360p & 720p & 720p & 720p & 720p & 720p & \cellcolor{DFOurs}480p \\
    Generation Horizon & Short & Medium & Medium & Medium & Long & Long & \cellcolor{DFOurs}Medium \\
    Motion control degree & Discrete & Discrete & Discrete & Continuous & Discrete & Discrete & \cellcolor{DFOurs}Discrete \\
    \bottomrule
  \end{tabular}}
  \caption{Feature-level comparison with recent interactive world models, compiled from public system descriptions and technical reports~\cite{matrixgame2,matrixgame3,worldplay,genie3,lingbotworld}. Checkmarks indicate reported support under non-uniform evaluation setups and do not imply equal quality, scale, or benchmark protocol. ``Diverse multimodal input'' refers to support for at least three input modalities: text, image, and video.}
  \label{tab:comparison}
\end{table}

\section{Related Work}

\textbf{Interactive and action-conditioned generation.} World models have long been framed as learned predictors of future states, but recent generative systems shift the interface toward controllable visual environments~\cite{ha2018worldmodels,genie2024}. GameNGen demonstrates real-time game-like visual generation with diffusion models, while Matrix-Game~2.0 formulates streaming world generation with frame-level keyboard/mouse conditioning and few-step autoregressive diffusion~\cite{gamengen2024,matrixgame2}. Matrix-Game~3.0 and WorldPlay extend this design space toward memory-augmented, real-time interactive rollout~\cite{matrixgame3,worldplay}.

\textbf{Open video priors and causal rollout.} Wan provides the open video-model foundation used in the LongLive lineage~\cite{wan2025}. LongLive extends a short-clip video model into frame-level autoregressive long-video generation using mechanisms such as KV recache, streaming long tuning, short-window attention, and frame sinks~\cite{longlive2025}. MAGI-1 and related autoregressive video systems further support the view that causal or chunked video generators can serve as substrates for streaming continuation~\cite{magi1,rollingforcing,ca2vdm}.

\textbf{Memory, multimodal entry, and promptability.} Persistent memory is now a defining frontier for interactive world simulation: WorldPlay, Infinite-World, DreamX-World, Genie~2, and Genie~3 all emphasize revisit consistency, off-screen persistence, or long-horizon context retention~\cite{worldplay,infiniteworld,dreamxworld,genie2,genie3}. Image-conditioned or promptable world entry points appear across Genie-style environments, Wan-family I2V generation, MAGI-style AR video generation, DreamX-World, WorldPlay, LongLive, and BiWM~\cite{genie2024,genie2,wan2025,magi1,dreamxworld,worldplay,longlive2025,biwm}. These systems define the context for \dfshort{}: broad preview-scale interaction with persistent spatial memory still left as the main missing capability.

\section{System Overview}

\dfshort{} is a preview-scale streaming autoregressive world model. A rollout begins from an initialization context and advances as a closed-loop visual process: future latent frames are predicted from visual history, text conditioning, and live \kmp{} actions; decoded observations are then recycled into the next prediction step. At the interface level, the preview combines prompt-only generation, image/video/mixed multimodal entry, live control, first- and third-person operation, mid-stream reprompting, and minute-scale continuation. The next section summarizes these observable capabilities before the report turns to the backbone adaptation, multimodal prefixing, runtime, and failure modes.

\section{Qualitative Capabilities}
\label{sec:capabilities}

\begin{figure}[H]
  \centering
  \includegraphics[width=\textwidth]{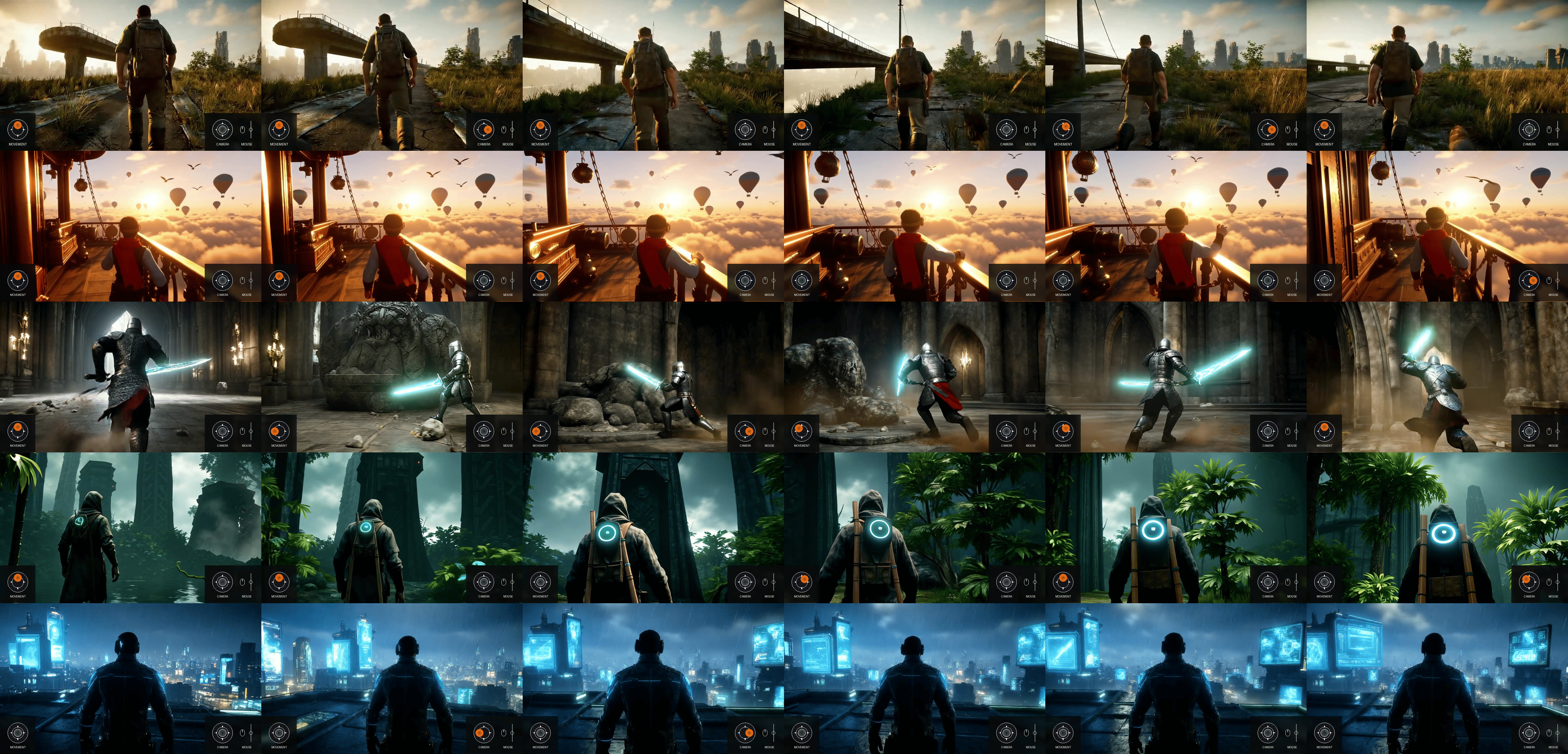}
  \caption{Representative third-person DF-World~0.1 Preview rollouts across multiple prompts and domains.}
  \label{fig:tpv_grid}
\end{figure}

The current preview exposes six user-visible capabilities in one runtime: prompt-only rollout, multimodal entry from text/image/video or mixed context, live keyboard and mouse control, first- and third-person view modes, mid-stream reprompting, and minute-scale continuation. These capabilities are not independent modules; they operate through the same autoregressive loop in which generated observations become the next conditioning history.

Prompt-only rollout provides the base interaction mode: a text description initializes a trajectory and the model advances the world as future visual observations. Multimodal entry extends the same loop by inserting image or video inputs into the latent history before continuation, allowing a session to begin from a provided visual state rather than from text alone. Once initialized, the rollout remains interactive: keyboard and mouse input condition the next generated frames, while the accumulated visual history supplies local temporal state.

Dual-view operation exposes two different control regimes. First-person mode emphasizes egocentric camera motion and navigation-like response, while third-person mode must coordinate character displacement, external camera behavior, parallax, and background continuation. The preview therefore uses separate view-specific action checkpoints rather than relying on a single shared controller to cover both observation mappings.

\begin{figure}[H]
  \centering
  \includegraphics[width=\textwidth]{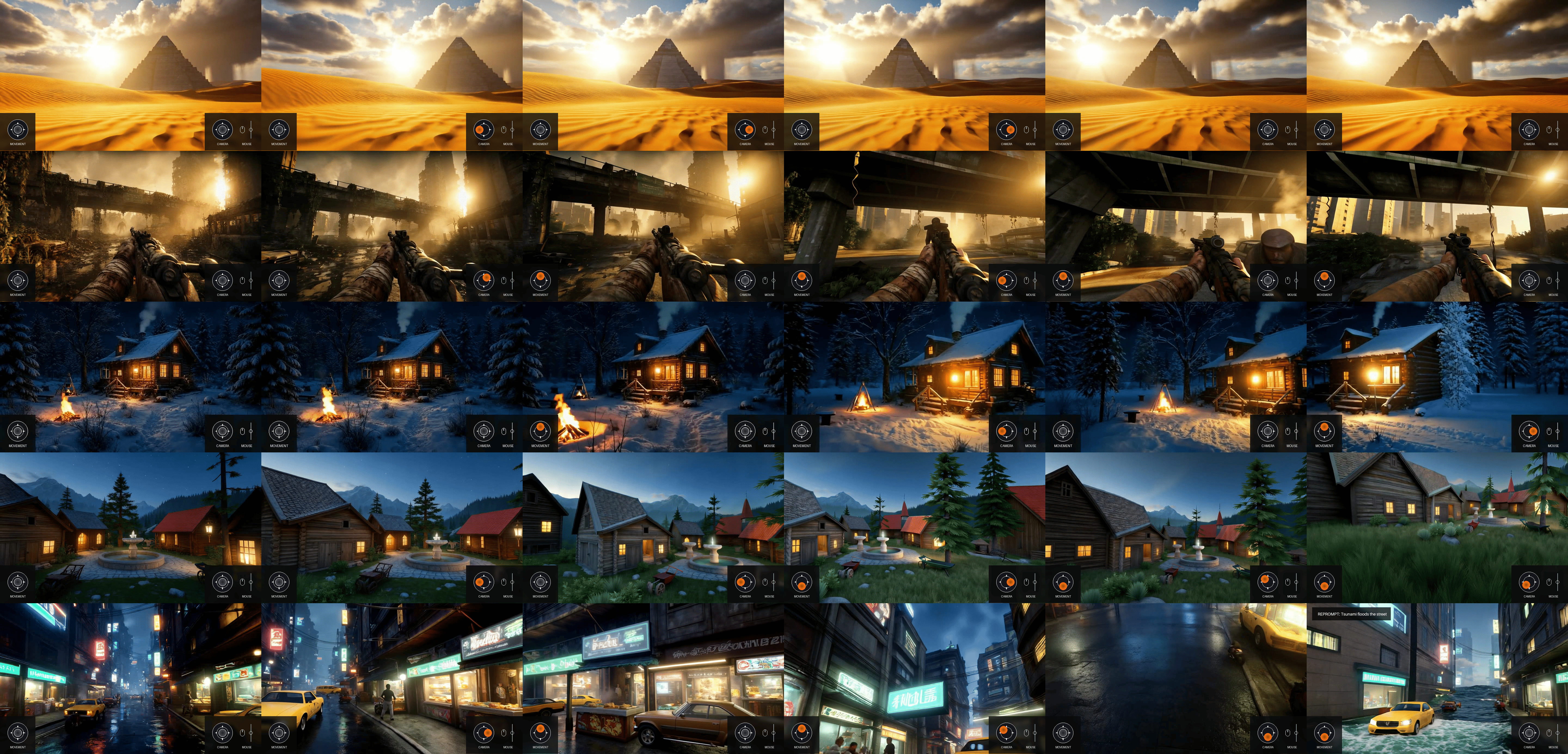}
  \caption{Representative first-person DF-World~0.1 Preview rollouts across multiple prompts and domains.}
  \label{fig:fpv_grid}
\end{figure}

\begin{figure}[H]
  \centering
  \includegraphics[width=\textwidth]{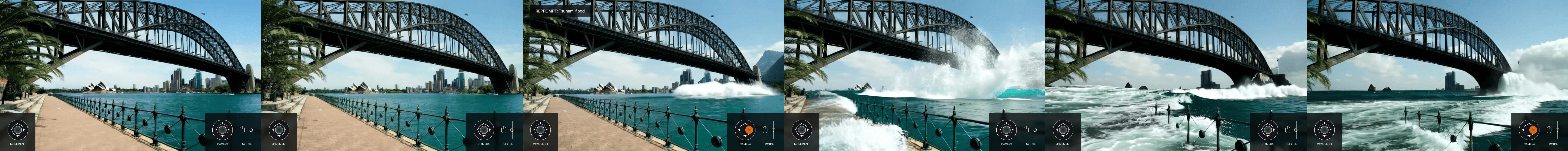}
  \caption{Mid-stream reprompting example. A running coastal rollout is redirected by a text update that summons a tsunami.}
  \label{fig:reprompt_strip}
\end{figure}

Mid-stream reprompting changes the semantic condition during an active rollout while retaining the current visual history. This enables promptable continuation and event injection without restarting the session, as in the coastal sequence in Figure~\ref{fig:reprompt_strip}.

\dfshort{} is not constrained to a fixed rollout horizon and can continue generation autoregressively; for the current preview, the first minute of inference provides the strongest qualitative window for interactive use.

\FloatBarrier

\section{Backbone, Control, and Adaptation}

\textbf{Backbone and rollout formulation.} \dfshort{} uses LongLive~1 as its causal autoregressive video substrate. LongLive~1 builds on Wan2.1-T2V-1.3B and provides the runtime properties retained in DF-World: frame-level rollout, prompt switching through KV recache, cacheable causal structure, short-window attention, frame-sink context, and efficient streaming inference~\cite{wan2025,longlive2025,longlivehf}. We retain this runtime as the base execution path and target DreamForge-side modifications at controllability, multimodal entry, and deployment behavior.

\textbf{Backbone LoRA.} We first train a rank-64 LoRA on top of the LongLive/Wan backbone without an action module. This stage adapts the video prior toward interactive game-like domains before introducing explicit user control. The training mix combines gameplay videos from the NitroGen dataset, GameGen-X/Open-World Video Game Dataset material, and a smaller manually collected high-quality set~\cite{nitrogen2026,gamegenx2024}. The LoRA is trained on both first-person and third-person clips; view selection is exposed through prompt conditioning, e.g. prompt tokens such as \emph{first person} or \emph{third person} activate the corresponding visual regime.

\textbf{Action dataset and labels.} The manually collected control subset contains 5-second clips at 24~FPS with continuous mouse motion and discrete movement/control annotations. We normalize and label interaction, jump, sprint, crouch, \texttt{act\_1} (RMB/L2), and \texttt{act\_2} (LMB/R2), together with view-type metadata across all data sources. Additional button channels are retained in the data pipeline for future training, while the present action-module adaptation uses the subset required by the first-person and third-person controllers.

\textbf{Residual action pathway.} After backbone LoRA training, we adapt Matrix-Game~2.0-style action conditioning to the LongLive runtime. Matrix-Game~2.0 reports a frame-level keyboard/mouse action injection module inside a few-step autoregressive diffusion loop, trained on roughly 1200 hours of interaction-annotated data~\cite{matrixgame2}. Given a strong pretrained prior, we transplant the Matrix-Game~2.0 action-module weights onto the LongLive-based DiT and then briefly fine-tune the transferred control path to reduce model-transfer error and align the action features with the adapted backbone.

\textbf{View-specific control checkpoints.} Dual-view control is implemented with two high-rank LoRA adaptations on top of the transferred Matrix-Game action module. The first LoRA is trained on first-person samples to establish stable egocentric control. The second is trained on third-person samples to support external-view control without forcing the same controller to cover two incompatible action-to-observation mappings. Training uses an $x_0$ MSE objective together with trajectory/pose losses; Depth Anything~3 is used as an auxiliary geometry source for pose/depth supervision during data preparation~\cite{depthanything3}. After training, the LoRA weights are fused into the original Matrix-Game action-module weights, producing two view-specific action checkpoints selected before inference according to the chosen view mode. This design provides dual-view control without redesigning the module and helps to avoid perspective drift between third-person and first-person regimes.

\section{Multimodal Initialization via Latent History Conditioning}

LongLive~1 is a text-to-video long-video system: its native interactive interface is prompt switching over autoregressive rollout, not image- or video-initialized world entry. \dfshort{} therefore adds multimodal initialization inside the DreamForge runtime. The implementation uses the autoregressive history interface, where images and videos are encoded into the same latent space used for rollout and inserted as clean initial history before continuation.

For image initialization, an input image is encoded by the video VAE into latent history and used as the initial observation prefix. The rollout continues from this prefix under the current text prompt and live action stream, producing an image-to-world entry point. For video initialization, a provided segment is encoded as multiple latent frames or chunks, after which the model predicts future observations autoregressively. Combined with LongLive-native prompt switching, this yields video-to-world or vid2vid-style continuation under user control.

Because the adopted LongLive~1 interface did not expose these entry modes when it was integrated as the DF-World backbone, the DreamForge runtime adds them through clean latent prefixes rather than a separate image-to-video initializer. Related systems expose image-conditioned or video-continuation interfaces through different architectures, including Genie-style playable worlds, Wan-family I2V/video tasks, MAGI-style chunked autoregressive video generation, WorldPlay continuation, and DreamX-World text/image-to-video generation~\cite{genie2024,genie2,wan2025,magi1,worldplay,dreamxworld}.

Prefix conditioning anchors the beginning of the rollout. Once the generated trajectory diverges from the initial context, persistence again depends on the autoregressive history.

\section{Runtime}

The inference stack is built around the LongLive~1 causal AR runtime because its prompt-switching, KV-cache, short-window attention, and frame-sink mechanisms already provide a strong foundation for streaming generation~\cite{longlive2025}. DreamForge modifies this runtime substantially for interactive deployment: we add asynchronous streaming around generation and VAE decoding, tune low-compute execution paths, implement KV-cache quantization, and integrate Deep Forcing-style training-free cache management to reduce rollout drift during long interactive sequences~\cite{deepforcing2025}. In addition to the default Wan2.1 VAE path, we benchmark a LightTAEW~2.1 decoding path from the LightX2V/ComfyUI-LightVAE release, which provides a faster lightweight TAE decoder for Wan-family workflows~\cite{lightvae2026}. The runtime also integrates the view-specific action-module checkpoints described above.

Runtime is measured as observed end-to-end preview throughput at native $480{\times}832$ resolution. Measurements include the practical generation path---diffusion transformer execution, action conditioning, VAE decoding, and streaming overhead---rather than isolated transformer-only throughput, and therefore reflect the frame rate observed through the DreamForge runtime.

\begin{table}[H]
  \centering
  \scriptsize
  \setlength{\tabcolsep}{2.3pt}
  \renewcommand{\arraystretch}{1.06}
  \begin{minipage}[t]{0.485\textwidth}
    \centering
    \textbf{Default VAE path}\vspace{0.25em}

    \rowcolors{2}{white}{DFTable}
    \begin{tabularx}{\linewidth}{@{}>{\raggedright\arraybackslash}X c c c@{}}
      \toprule
      \textbf{Hardware} & \textbf{Prec.} & \textbf{Runtime} & \textbf{VRAM} \\
      \midrule
      RTX 4090 & bf16 & $\sim$10 FPS & $\sim$9GB \\
      RTX 4090 & fp8 & $\sim$12 FPS & $\sim$5GB \\
      H100 & bf16 & $\sim$15 FPS & $\sim$9GB \\
      H100 & fp8 & $\sim$17 FPS & $\sim$5GB \\
      \bottomrule
    \end{tabularx}
  \end{minipage}\hfill
  \begin{minipage}[t]{0.485\textwidth}
    \centering
    \textbf{LightTAEW~2.1 path}\vspace{0.25em}

    \rowcolors{2}{white}{DFTable}
    \begin{tabularx}{\linewidth}{@{}>{\raggedright\arraybackslash}X c c c@{}}
      \toprule
      \textbf{Hardware} & \textbf{Prec.} & \textbf{Runtime} & \textbf{VRAM} \\
      \midrule
      RTX 4090 & bf16 & $\sim$12 FPS & $\sim$8GB \\
      RTX 4090 & fp8 & 14--15 FPS & $\sim$4GB \\
      H100 & bf16 & $\sim$18 FPS & $\sim$8GB \\
      H100 & fp8 & $\sim$19 FPS & $\sim$4GB \\
      \bottomrule
    \end{tabularx}
  \end{minipage}
  \caption{Preview runtime measurements at 480$\times$832. The left table reports the default Wan2.1 VAE path; the right table reports the same runtime with LightTAEW~2.1 decoding~\cite{lightvae2026}.}
  \label{tab:runtime}
\end{table}

LightTAEW~2.1 improves the measured preview throughput and memory footprint, while the default VAE path remains the reference path for quality-preserving demonstrations. LongLive~2.0 and related systems show the same broader direction--quantized caches, asynchronous decoding, and low-precision runtime infrastructure--but \dfshort{} uses its own preview-specific implementation choices~\cite{longlive2}.

\section{Limitations}
\label{sec:limitations}

\begin{figure}[H]
  \centering
  \includegraphics[width=\textwidth]{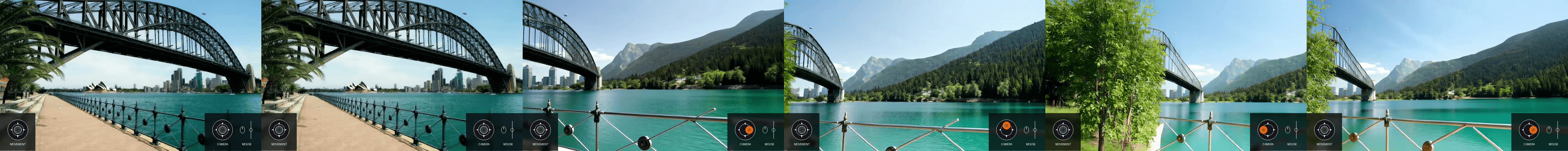}
  \caption{Revisit-consistency failure: rotating away and returning introduces previously unseen trees and scene structure.}
  \label{fig:memory_limitation}
\end{figure}

\textbf{Memory and revisit consistency.} Persistent spatial memory is the dominant missing capability. \dfshort{} can continue a trajectory, but it does not maintain a reliable external map of generated space. When the user rotates away from a region and later returns, scene content may be re-synthesized rather than preserved. Figure~\ref{fig:memory_limitation} shows a representative absence-of-memory case: after camera rotation and return, new trees and scene structure appear in a location that was not previously present. This places the preview behind memory-augmented systems that explicitly target revisit consistency through retrieval, context reconstruction, hierarchical memory, or camera-aware conditioning~\cite{matrixgame3,worldplay,infiniteworld,dreamxworld,genie2,genie3}.

\begin{figure}[H]
  \centering
  \includegraphics[width=\textwidth]{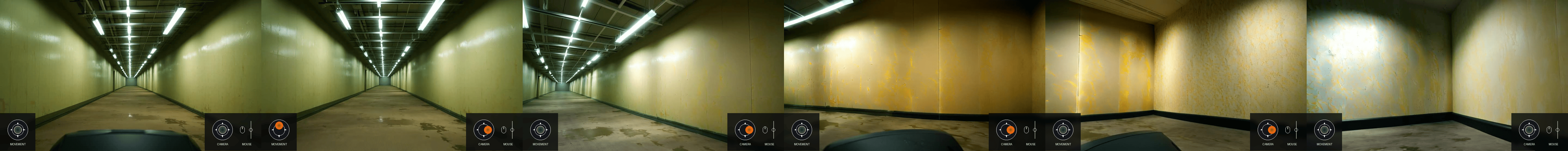}
  \caption{Color and texture degradation during self-conditioned rollout. Later frames partially restore color balance, but fine texture detail remains degraded.}
  \label{fig:degradation_limitation}
\end{figure}

\textbf{Long-horizon drift.} The system supports minute-scale interactive rollouts, but quality, object identity, and layout consistency degrade as the model conditions on its own imperfect history. Figure~\ref{fig:degradation_limitation} shows a long self-conditioned rollout with visible color oversaturation and texture degradation; later frames partially recover color balance, but not the lost texture fidelity.

\textbf{Control, latency, and sensory scope.} Keyboard and mouse actions influence the trajectory, but action diversity and precision remain below explicit simulation. Aggressive camera motion can destabilize the scene, and third-person control is more fragile than first-person navigation. Latency is real-time, but not yet at conventional game-feel responsiveness. The preview does not include sound generation, multi-agent interaction, or reliable physical interaction with persistent objects.

\section{Adaptation Scale}

\dfshort{} uses 64 hours of curated gameplay/control video across the backbone LoRA and two action-module LoRA stages. This refers to the filtered adaptation set used for the preview system, not the raw size of the upstream corpora used during data sourcing.

Public reports do not expose a uniform cost metric for interactive world models. We therefore compare reported training-video scale where direct hour-level figures are available from primary sources.~\cite{matrixgame2,genie2024}.
\begin{figure}[H]
  \centering
  \includegraphics[width=0.72\textwidth]{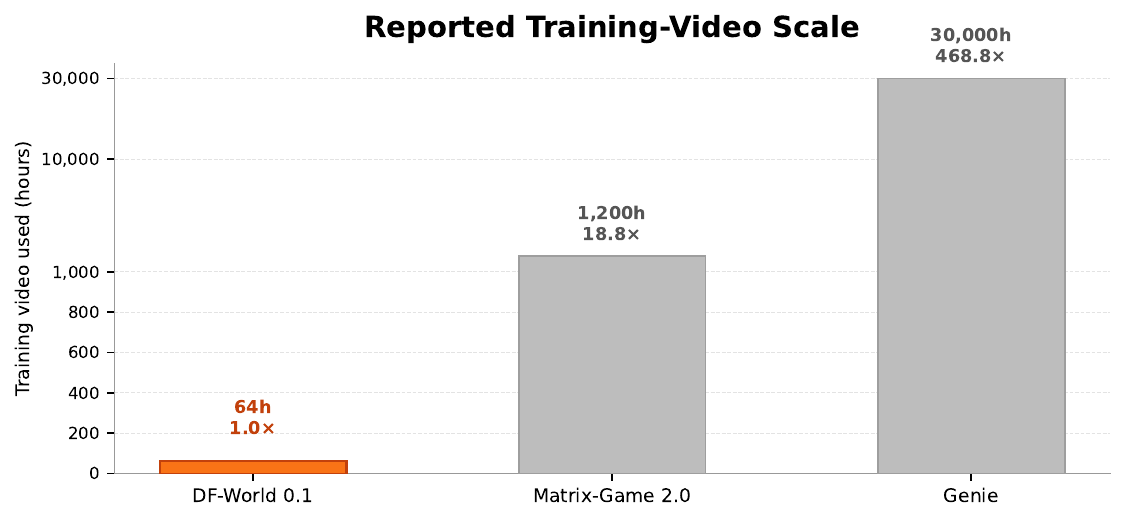}
  \caption{Reported training-video scale for systems with public hour-level training reports. Bars are annotated with raw hours and scale relative to DF-World's 64-hour curated adaptation set. For readability, the visual scale is softly compressed above 1{,}000 hours while preserving the ordering and large separation between systems. Matrix-Game~2.0 reports about 1200 hours of interaction-annotated data; Genie reports a 30{,}000-hour filtered platformer training set~\cite{matrixgame2,genie2024}.}
  \label{fig:training_scale}
\end{figure}
\section{Release Status and Next Steps}

\dfshort{} is a preview, not a full model release. DreamForge does not currently open-source the preview checkpoint. DF-World~0.5 is planned as a more complete model generation and may be released publicly if licensing, safety, and engineering constraints allow.

The next technical target is persistent spatial memory. Candidate directions include latent memory retrieval, camera-aware memory, loop-closure training, self-generated-history training, and hybrid external scene-state mechanisms. Beyond memory, we are exploring larger video backbones, stronger action-module architectures, improved dual-view control, streaming multimodal generation in which image and video inputs can be updated continuously rather than only used to seed real-time video output, and early audiovisual real-time world generation with synchronized audio conditioned on the generated visual state. Additional work is needed on latency, action diversity, third-person stability, aggressive-camera robustness, and systematic evaluation.

\section{Conclusion}

\dfshort{} demonstrates a constrained-compute path toward real-time controllable world models. By adapting a LongLive/Wan video prior, integrating a Matrix-Game-style residual action pathway, exposing multimodal initialization through latent history conditioning, and optimizing the inference stack for consumer GPUs, DreamForge obtains a live generated-world loop supporting text/image/video initialization, mid-stream reprompting, and dual-view control.

Persistent spatial memory, frontier-scale fidelity, and precise physical control remain open. The preview's technical relevance lies in its operating point: broad interactive capability coverage, single-GPU runtime, and targeted adaptation cost. Future work will focus on persistent memory, lower latency, larger backbones, stronger action conditioning, streaming multimodal and audiovisual control, and more stable dual-view interaction.

\section*{Acknowledgments}

DreamForge AI Lab is self-funded. The authors thank the LongLive/NVIDIA Labs team for releasing a strong causal autoregressive long-video foundation for this line of work, and the Matrix-Game team for open action-conditioned world-model research that informed the control pathway used in this preview. We also thank the broader open video generation and world-modeling communities whose released models, papers, and implementation details made constrained-compute experimentation possible.


\begin{thebibliography}{99}

\bibitem{ha2018worldmodels}
David Ha and Jürgen Schmidhuber.
\newblock World Models.
\newblock \href{https://arxiv.org/abs/1803.10122}{arXiv:1803.10122}, 2018.

\bibitem{genie2024}
Jake Bruce, Michael Dennis, Ashley Edwards, Jack Parker-Holder, Yuge Shi, Edward Hughes, Matthew Lai, Aditi Mavalankar, Richie Steigerwald, Chris Apps, Yusuf Aytar, Sarah Bechtle, Feryal Behbahani, Stephanie Chan, Nicolas Heess, Lucy Gonzalez, Simon Osindero, Sherjil Ozair, Scott Reed, Jingwei Zhang, Konrad Zolna, Jeff Clune, Nando de Freitas, Satinder Singh, and Tim Rocktäschel.
\newblock Genie: Generative Interactive Environments.
\newblock \href{https://arxiv.org/abs/2402.15391}{arXiv:2402.15391}, 2024.

\bibitem{genie2}
Google DeepMind.
\newblock Genie 2: A large-scale foundation world model.
\newblock Technical announcement, 2024. \href{https://deepmind.google/blog/genie-2-a-large-scale-foundation-world-model/}{deepmind.google/blog/genie-2-a-large-scale-foundation-world-model}.

\bibitem{genie3}
Google DeepMind.
\newblock Genie 3: A new frontier for world models.
\newblock Technical announcement, 2025. \href{https://deepmind.google/blog/genie-3-a-new-frontier-for-world-models/}{deepmind.google/blog/genie-3-a-new-frontier-for-world-models}.

\bibitem{gamengen2024}
Dani Valevski et al.
\newblock Diffusion Models Are Real-Time Game Engines.
\newblock \href{https://arxiv.org/abs/2408.14837}{arXiv:2408.14837}, 2024.

\bibitem{wan2025}
Wan Team.
\newblock Wan: Open and Advanced Large-Scale Video Generative Models.
\newblock \href{https://arxiv.org/abs/2503.20314}{arXiv:2503.20314}, 2025. Code and models: \href{https://github.com/Wan-Video/Wan2.1}{github.com/Wan-Video/Wan2.1}.

\bibitem{longlive2025}
Shuai Yang, Wei Huang, Ruihang Chu, Yicheng Xiao, Yuyang Zhao, Xianbang Wang, Muyang Li, Enze Xie, Yingcong Chen, Yao Lu, Song Han, and Yukang Chen.
\newblock LongLive: Real-time Interactive Long Video Generation.
\newblock \href{https://arxiv.org/abs/2509.22622}{arXiv:2509.22622}, 2025.

\bibitem{longlivehf}
LongLive Team.
\newblock LongLive-1.3B model release materials.
\newblock Hugging Face paper page, 2025. \href{https://huggingface.co/papers/2509.22622}{huggingface.co/papers/2509.22622}.

\bibitem{matrixgame2025}
Yifan Zhang et al.
\newblock Matrix-Game: Interactive World Foundation Model.
\newblock \href{https://arxiv.org/abs/2506.18701}{arXiv:2506.18701}, 2025.

\bibitem{matrixgame2}
Xianglong He, Chunli Peng, Zexiang Liu, Boyang Wang, Yifan Zhang, Qi Cui, Fei Kang, Biao Jiang, Mengyin An, Yangyang Ren, Baixin Xu, Hao-Xiang Guo, Kaixiong Gong, Cyrus Wu, Wei Li, Xuchen Song, Yang Liu, Eric Li, and Yahui Zhou.
\newblock Matrix-Game 2.0: An Open-Source, Real-Time, and Streaming Interactive World Model.
\newblock \href{https://arxiv.org/abs/2508.13009}{arXiv:2508.13009}, 2025.

\bibitem{matrixgame3}
Zile Wang, Zexiang Liu, Jaixing Li, Kaichen Huang, Baixin Xu, Fei Kang, Mengyin An, Peiyu Wang, Biao Jiang, Yichen Wei, Yidan Xietian, Jiangbo Pei, Liang Hu, Boyi Jiang, Hua Xue, Zidong Wang, Haofeng Sun, Wei Li, Wanli Ouyang, Xianglong He, Yang Liu, Yangguang Li, and Yahui Zhou.
\newblock Matrix-Game 3.0: Real-Time and Streaming Interactive World Model with Long-Horizon Memory.
\newblock \href{https://arxiv.org/abs/2604.08995}{arXiv:2604.08995}, 2026.

\bibitem{worldplay}
Wenqiang Sun, Haiyu Zhang, Haoyuan Wang, Junta Wu, Zehan Wang, Zhenwei Wang, Yunhong Wang, Jun Zhang, Tengfei Wang, and Chunchao Guo.
\newblock WorldPlay: Towards Long-Term Geometric Consistency for Real-Time Interactive World Modeling.
\newblock \href{https://arxiv.org/abs/2512.14614}{arXiv:2512.14614}, 2025. Project page: \href{https://3d-models.hunyuan.tencent.com/world/}{3d-models.hunyuan.tencent.com/world}.

\bibitem{infiniteworld}
Ruiqi Wu, Xuanhua He, Meng Cheng, Tianyu Yang, Yong Zhang, Zhuoliang Kang, Xunliang Cai, Xiaoming Wei, Chunle Guo, Chongyi Li, and Ming-Ming Cheng.
\newblock Infinite-World: Scaling Interactive World Models to 1000-Frame Horizons via Pose-Free Hierarchical Memory.
\newblock \href{https://arxiv.org/abs/2602.02393}{arXiv:2602.02393}, 2026.

\bibitem{dreamxworld}
DreamX Team.
\newblock DreamX-World 1.0: A General-Purpose Interactive World Model.
\newblock \href{https://arxiv.org/abs/2606.16993}{arXiv:2606.16993}, 2026.

\bibitem{lingbotworld}
Robbyant Team.
\newblock LingBot-World: Advancing Open-source World Models.
\newblock \href{https://arxiv.org/abs/2601.20540}{arXiv:2601.20540}, 2026.

\bibitem{magi1}
Sand.ai Team.
\newblock MAGI-1: Autoregressive Video Generation at Scale.
\newblock \href{https://arxiv.org/abs/2505.13211}{arXiv:2505.13211}, 2025. Code: \href{https://github.com/SandAI-org/MAGI-1}{github.com/SandAI-org/MAGI-1}.

\bibitem{rollingforcing}
Kunhao Liu, Wenbo Hu, Jiale Xu, Ying Shan, and Shijian Lu.
\newblock Rolling Forcing: Autoregressive Long Video Diffusion in Real Time.
\newblock \href{https://arxiv.org/abs/2509.25161}{arXiv:2509.25161}, 2025.

\bibitem{ca2vdm}
Kaifeng Gao, Jiaxin Shi, Hanwang Zhang, Chunping Wang, Jun Xiao, and Long Chen.
\newblock Ca2-VDM: Efficient Autoregressive Video Diffusion Model with Causal Generation and Cache Sharing.
\newblock \href{https://arxiv.org/abs/2411.16375}{arXiv:2411.16375}, 2024. Code: \href{https://github.com/Dawn-LX/CausalCache-VDM}{github.com/Dawn-LX/CausalCache-VDM}.

\bibitem{nitrogen2026}
Loic Magne, Anas Awadalla, Guanzhi Wang, Yinzhen Xu, Joshua Belofsky, Fengyuan Hu, Joohwan Kim, Ludwig Schmidt, Georgia Gkioxari, Jan Kautz, Yisong Yue, Yejin Choi, Yuke Zhu, and Linxi Fan.
\newblock NitroGen: An Open Foundation Model for Generalist Gaming Agents.
\newblock \href{https://arxiv.org/abs/2601.02427}{arXiv:2601.02427}, 2026.

\bibitem{gamegenx2024}
Haoxuan Che, Xuanhua He, Quande Liu, Cheng Jin, and Hao Chen.
\newblock GameGen-X: Interactive Open-world Game Video Generation.
\newblock \href{https://arxiv.org/abs/2411.00769}{arXiv:2411.00769}, 2024.

\bibitem{depthanything3}
Haotong Lin, Sili Chen, Junhao Liew, Donny Y. Chen, Zhenyu Li, Guang Shi, Jiashi Feng, and Bingyi Kang.
\newblock Depth Anything 3: Recovering the Visual Space from Any Views.
\newblock \href{https://arxiv.org/abs/2511.10647}{arXiv:2511.10647}, 2025.

\bibitem{deepforcing2025}
Jung Yi, Wooseok Jang, Paul Hyunbin Cho, Jisu Nam, Heeji Yoon, and Seungryong Kim.
\newblock Deep Forcing: Training-Free Long Video Generation with Deep Sink and Participative Compression.
\newblock \href{https://arxiv.org/abs/2512.05081}{arXiv:2512.05081}, 2025.

\bibitem{lightvae2026}
ModelTC.
\newblock ComfyUI-LightVAE: High-Performance VAE Custom Nodes for LightX2V, including LightVAE and LightTAE models.
\newblock GitHub repository, 2026. \href{https://github.com/ModelTC/ComfyUI-LightVAE}{github.com/ModelTC/ComfyUI-LightVAE}.

\bibitem{longlive2}
Yukang Chen, Luozhou Wang, Wei Huang, Shuai Yang, Bohan Zhang, Yicheng Xiao, Ruihang Chu, Weian Mao, Qixin Hu, Shaoteng Liu, Yuyang Zhao, Huizi Mao, Ying-Cong Chen, Enze Xie, Xiaojuan Qi, Song Han.
\newblock LongLive-2.0: An NVFP4 Parallel Infrastructure for Long Video Generation.
\newblock \href{https://arxiv.org/abs/2605.18739}{arXiv:2605.18739}, 2026.

\bibitem{biwm}
Shaohao Rui, Xiaofeng Mao, Zhanyu Zhang, Peijia Lin, Yansong Zhu, Yibo Zhang, Haibin Wan, and Weijie Ma.
\newblock BiWM: Advancing Open-Source Interactive Video World Models with Bidirectional Autoregression.
\newblock \href{https://arxiv.org/abs/2606.10135}{arXiv:2606.10135}, 2026.

\end{thebibliography}
\end{document}